\documentclass[conference]{IEEEtran}
\IEEEoverridecommandlockouts
\usepackage{footnote}
\usepackage{cite}
\usepackage{amsmath,amssymb,amsfonts}
\usepackage{algorithmic}
\usepackage{graphicx}
\usepackage{textcomp}
\usepackage{xcolor}
\def\BibTeX{{\rm B\kern-.05em{\sc i\kern-.025em b}\kern-.08em
    T\kern-.1667em\lower.7ex\hbox{E}\kern-.125emX}}

\usepackage{times}
\usepackage{latexsym}

\usepackage[T1]{fontenc}

\usepackage[utf8]{inputenc}

\usepackage{microtype}

\usepackage{inconsolata}

\usepackage{graphicx}
\usepackage{longtable}
\usepackage{listings}
\usepackage{hyperref}
\title{Execution-Based Evaluation of Natural Language to Bash and PowerShell for Incident Remediation}

\author{Ngoc Phuoc An Vo, Brent Paulovicks, Vadim Sheinin \\
  IBM Research, New York, USA \\
  {\texttt{ngoc.phuoc.an.vo@ibm.com}, \texttt{ovicks@us.ibm.com}, \texttt{vadims@us.ibm.com}}}

\begin{document}
\maketitle
\begin{abstract}
Given recent advancements of Large Language Models (LLMs), code generation tasks attract immense attention for wide application in different domains. In an effort to evaluate and select a best model to automatically remediate system incidents discovered by Application Performance Monitoring (APM) platforms, it is crucial to verify if the generated code is syntactically and semantically correct, and whether it can be executed correctly as intended. However, current methods for evaluating the quality of code generated by LLMs heavily rely on surface form similarity metrics (e.g. BLEU, ROUGE, and exact/partial match) which have numerous limitations.
In contrast, execution based evaluation focuses more on code functionality and does not constrain the code generation to any fixed solution. Nevertheless, designing and implementing such execution-based evaluation platform is not a trivial task. There are several works creating execution-based evaluation platforms for popular programming languages such as SQL, Python, Java, but limited or no attempts for scripting languages such as Bash and PowerShell. In this paper, we present the first execution-based evaluation platform in which we created three test suites (total 125 hand-crafted test cases) to evaluate Bash (both single-line commands and multiple-line scripts) and PowerShell codes generated by LLMs. We benchmark seven closed and open-source LLMs using our platform with different techniques (zero-shot vs. few-shot learning). 



\end{abstract}


\section{Introduction}

Code generation is a task of generating a target programming language from a prompt in natural language. Among popular programming languages, Bash is a Unix shell and command language widely used in GNU/Linux Operating System for automating different tasks, such as performance monitoring, compilation, system administration, system diagnostics, etc. Meanwhile, PowerShell is a cross-platform task automation solution made up of a command-line shell, a scripting language, and a configuration management framework running on Windows, Linux, and macOS. The task NL2Bash was first introduced in \cite{b1} to translate from natural language descriptions to Bash commands. Recent works also attempt to use LLMs to automatically generate PowerShell for various purposes \cite{b9, b10, b8}.

An important step of code generation is verifying the generated code if it is syntactically and semantically correct, and whether it can be executed correctly as intended. Especially code verification is crucial for Application Performance Monitoring (APM) platforms (e.g. Instana, AppDynamics) to automate remediation actions. There are different approaches for assessing code quality such as manual verification by a human or automatic verification using various evaluation metrics. However, current methods for automatically evaluating code quality heavily rely on surface form similarity metrics such as BLEU, ROUGE, and exact/partial match. This may overlook code syntax features, ignore semantics features and execution effects, or even over-penalize alternative solutions \cite{b11}. In contrast, execution-based evaluation focuses more on code functionality and does not constrain the code generation to any fixed solution. Nevertheless, execution-based evaluation, especially for Bash and PowerShell is rather complex and costly to design and implement since the verification should take many factors (e.g. changes of test environment) into account. In light of automating remediation actions for incidents discovered from various platforms in APM tools, the main idea is to create an execution-based evaluation dataset and platform for Bash and PowerShell codes to select the best code model for automatic incident remediation.


\section{Related Works}
Quality assessment for code generation tasks is verifying and testing the generated code to ensure its correctness and quality. Execution-based evaluation has been proposed and widely used for SQL \cite{b2} (e.g. translating text to SQL) or logical forms \cite{b3} (e.g. translating language to logical forms). Some recent works on execution-based evaluation \cite{b4,b5,b6,b7} focus more on popular programming languages like Python. Another work \cite{b11} attempted to create an interactive environment for code assessment including Bash, however, it reused a subset of 200 prompts from the public dataset NL2Bash \cite{b1} to evaluate only single-line Bash commands. In contrast to this work, to the best of our knowledge, our work is the first one to create an execution-based evaluation platform for Bash (both single-line commands and multiple-line scripts) and PowerShell.

\section{Execution-based Evaluation Platform}
\begin{figure}[htbp]
\centerline{\includegraphics[scale=0.57]{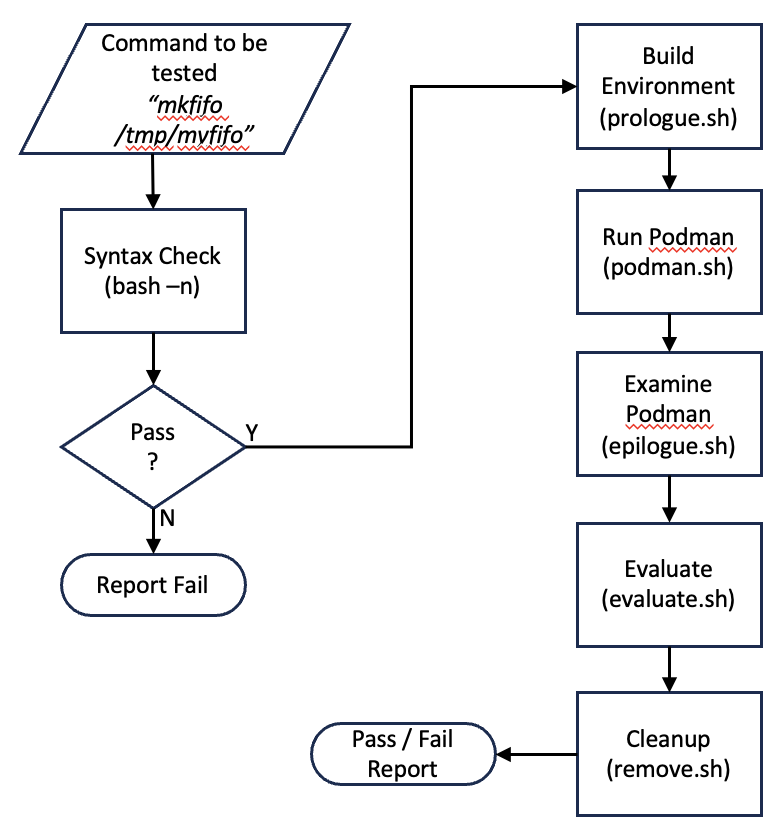}}
\caption{An Execution-based Evaluation Platform.}
\label{fig-overview}
\end{figure}

Execution-based evaluation focuses more on code functionality and does 
not constrain the code generation to any fixed solution. Given a predicted code, we execute it and compare its actual result against the expected result. 
Figure \ref{fig-overview} shows the overview of our execution-based evaluation platform consisting of different steps from building an environment to running and examining a container, and finally evaluating the result of executing such given code.

\subsection{Build Environment}

In this step, we build an environment to run our test cases.
Each test has a separate directory structure containing everything necessary to create the environment for a podman container to test a model's response to the prompt. Each test has a \texttt{name} file containing a short description of the test used only for labelling output; a \texttt{prompt} file containing the prompt used to elicit the response (code) to be tested; a \texttt{bash.sh} file containing a command which satisfies the prompt and is used as a sanity check verifying the entire test environment whenever changes are made.
 
A \texttt{home/test} directory is the user test's prototype home directory in the container environment. This directory will be copied to a working directory under the \texttt{log} directory before the container is run and discarded afterwards providing a clean initial setup for each test run. It contains \texttt{test.sh}, the test driver which will initialize the container environment via \texttt{pre\_test.sh}, run the model's response contained in \texttt{bash.sh}, and then collect any test output with \texttt{post\_test.sh}. For some tests, it may be necessary to include ancillary executables which are placed in \texttt{home/test/bin}.
 
Building the container environment for every test starts by running \texttt{prologue.sh}, a script which copies \texttt{home} from \texttt{tests/testNNN} to the working directory \texttt{log/testNNN}. It also looks for a test-specific \texttt{prologue.sh} and, if it exists, executes it to perform any additional setup required before the container is run.  The final step in \texttt{prologue.sh} is to create \texttt{log/testNNN/home/test/bash.sh} with the model's response to the prompt.


\subsection{Run Container}
The \texttt{podman.sh} script checks for the file \texttt{podman.opts} which is used for any additional podman options required for this particular test such as adding network access or mounting an additional file system. The following options are used for every test: the host user \texttt{test} user ID and group ID are mapped into the container's namespace using \texttt{--userns} which makes it easier to access files inside and outside the container. The home directory, \texttt{log/testNNN/home}, is mounted on \texttt{/home} inside the container using \texttt{--mount} and set to be the working directory via \texttt{--workdir}. The \texttt{stdout} and \texttt{stderr} output from podman is sent to \texttt{podman.log} in \texttt{log/testNNN} with \texttt{--log} and, finally, a limit is placed on the container's runtime with \texttt{--timeout}.
 
Podman always runs in "rootless" mode to minimize the access to the host filesystem. User \texttt{test}, after all, has sudo privileges in order to run some commands and, on occasion, a model has suggested \texttt{userdel} or \texttt{rm -fr} as a solution. The "rootless" mode also has the advantage of keeping container names local to the host user avoiding collisions with other users on the system.
 
The podman container uses RedHat's "universal base image" \texttt{ubi-init} (multi-service) with just a few additions to create user \texttt{test}, add them to the \texttt{wheel} group, and to change \texttt{wheel} to use passwordless sudo. All tests use this one common image which is built only once and not recreated for every test run.


\subsection{Inside The Container}
The Appendix \ref{subsec:verifier} shows an example of the sequence that happens inside the container. Once it starts running, there is usually a need to continue with initialization. The script \texttt{test.sh} runs \texttt{pre\_test.sh} to do this further setup. Examples of this are creating a dummy user when testing the prompt requesting help with the \texttt{userdel} command or to edit \texttt{/var/utmp} for any prompt soliciting user information via the \texttt{who} or \texttt{w} commands since \texttt{/var/utmp} is normally empty inside the container.
 
The \texttt{test.sh} script, itself, can vary from test to test. It was decided that the model would not be penalized for not including \texttt{sudo} in any command requiring superuser access so, in those cases, \texttt{test.sh} includes \texttt{sudo bash.sh}.
 
Another case is when the prompt is expected to generate a shell variable so \texttt{test.sh} executes the \texttt{bash.sh} in its current context \texttt{"bash.sh"} so \texttt{post\_test.sh} can test for the variables. The default is for \texttt{test.sh} to run \texttt{bash.sh} in a subshell \texttt{(bash.sh)}; this was learned the hard way when a model generated an \texttt{exit} command which prevented \texttt{post\_test.sh} from running.
 
Finally, \texttt{post\_test.sh} is executed by \texttt{test.sh} to extract information which can only be accessed from within the container, e.g., the date/time for a prompt which should have generated a change to the date/time or a change in the timezone. It also places some information in the \texttt{stdout} stream using the format \texttt{\#++VAR=something} for the evaluation step that runs outside the container.
 
The script \texttt{post\_test.sh} will sometimes output information to \texttt{stdout} which make it easier for a person reading the log file to understand what did or did not happen, e.g., the date before and after a timezone change.

\subsection{Examine Container}
This is very test-specific due to the variety of the commands being tested and the near-infinite number of ways they can go wrong. After podman exits, \texttt{epilogues.sh} runs the podman \texttt{diff} command which gives a nice summary of added, changed, and deleted files and directories. Unfortunately, podman does not track changes to mounted filesystems so a script was created to log the changes to the \texttt{home} directory.
 
Since some filesystem changes may be expected, for example, adding a new user touches a number of files (\texttt{/etc/passwd, /etc/group, /etc/shadow, /etc/gshadow, /etc/subuid, etc}), the general \texttt{epilogue.sh} script looks for a test-specific \texttt{epilogue.sh} in \texttt{tests/testNNN} which is used to ignore (remove) any expected filesystem changes from the log.


\subsection{Evaluate}
The evaluation script, \texttt{evaluate.sh}, starts by pulling the \texttt{stdout} and \texttt{stderr} outputs from \texttt{podman.log}. Any output on \texttt{stderr} is considered a failure, although some commands normally output to \texttt{stderr} so this has to be anticipated.
 
Any filesystem changes which were not expected and removed from the logs by the test-specific \texttt{epilogue.sh} are considered a failure. Conversely, any expected change to the filesystem which did not occur is also a failure.
 
The rest of the evaluation is specific to the test. Variables which were passed by \texttt{post\_test.sh} via \texttt{stdout} using
\texttt{\#++VAR=something} are extracted and their values tested for correctness and, in general, \texttt{grep} is used to scan \texttt{stdout} for strings which should appear and for strings which should not appear.

\subsection{Cleanup}
Finally, the cleanup script, \texttt{remove.sh}, is run to delete the working directory \texttt{log/testNNN}, and the podman container. Fortunately, such cleanup has been easy, i.e., no problems with a container hanging.  One minor problem which does arise occasionally is a file created by \texttt{root} \textit{inside} the container as that file's \texttt{uid/gid} are mapped to the test user's \texttt{subuid} and \texttt{subgid} namespace \textit{outside} the container and the \texttt{test} user does not have permission to remove the file nor does it have \texttt{sudo} access outside of the container.

\section{New Dataset and Experiments}
\textbf{Dataset Creation:} IBM Instana is an application performance monitoring (APM) tool helping users monitor and optimize performance of their software applications. It provides real-time insights into the health, availability, and performance of applications and infrastructure. We collected thousands of incidents discovered by Instana that can be resolved by different programming languages (Python, Ansible, Bash, PowerShell, etc) in various platforms. We filtered and grouped them into different categories and tasks for Bash and PowerShell such as system diagnostics, system administration, performance monitoring, system remediation, logic reasoning, math calculations, etc. The main idea is to create an execution-based evaluation dataset and platform for Bash and PowerShell codes to select the best code model to automatically remediate incidents discovered from various platforms in APM tools. We worked with our Site Reliability Engineers (SREs) to hand-craft 125 test cases (prompts and verifiers) for execution-based evaluation for Bash and PowerShell organized in 3 test suites: Single-line Bash commands (50 test cases), Multiple-line Bash scripts (50 test cases), and PowerShell (25 test cases). Our execution-based evaluation dataset and platform will be released for public access soon.

\textbf{Experiments:} We use our three test suites to benchmark the latest closed-source model GPT-4o \cite{b18} and six open-source models from HuggingFace: CodeLlama-34b-instruct \cite{b14}, Mistral-7B-Instruct-v0.1 \cite{b15}, Mixtral-8x7B-Instruct-v0.1 \cite{b16}, Granite-20b-code-instruct, Granite-34b-code-instruct, and Granite-34b-code-base \cite{b17}. We evaluate these models using zero-shot, 5-shot, and 10-shot settings.

For zero-shot experiments, we consider between Question/Answer (Q/A) prompt format (e.g. \textit{What is the bash command to ...?}) and imperative prompt format (e.g. \textit{Write a bash command to ...?}). We chose the Q/A format since it returns slightly better performance than imperative counterpart. All prompts are self-contained.

For few-shot learning, we randomly select 5 and 10 examples from independent datasets as follows:
\begin{itemize}
    \item Single-line Bash: NL2Bash \cite{b1}
    \item  Multiple-line Bash: we crawled a small set of Bash scripts from internet.
    \item PowerShell: Mega Collection repo.\footnote{https://github.com/fleschutz/PowerShell}
\end{itemize}

\textbf{Post-processing:} since different model outputs may have heterogeneous formats such as multiple suggested code snippets, nested code with descriptions or explanations, or even new pairs of questions and answers generated by models (hallucination), we implemented a heuristic algorithm for automatically extracting the first code occurrence/block from model predictions, the rest is discarded. 

\textbf{Evaluations:} As expected, GPT-4o outperforms other models across test settings (Table \ref{tab:evaluations}). In contrast, other six open-source LLMs show that:
\begin{itemize}
    \item For Single-line Bash, most models perform well with zero-shot. However, the results are contradictory for few-shot in which Granite-family models  improved in performance whereas other models declined.
    \item Since it is challenging to generate correct multiple-line Bash scripts, most models failed to achieve average accuracy in zero-shot. Noticeably, few-shot learning is not effective for any model.
    \item Most models achieve average accuracy for PowerShell in zero-shot and few-shot is helpful for most cases.
\end{itemize}
Except GPT-4o, results show that multiple-line Bash remains a challenging task for other LLMs. For the other two tasks, zero-shot achieves reasonable performance and few-shot contributes some improvement for most models. In general, Granite-family models achieve better performance or improvement in most cases.

Since most models perform poorly on multiple-line Bash scripts, we notice that there are 9 out of 50 test cases asking about system administration, the rest is math problems, hence, we investigate which task these models can do better. Table \ref{tab:evaluations_analysis_bash_scripts} shows that all models perform better on system administration tasks than math problems. 

\begin{table*}[htbp]
\caption{Execution-based Accuracy for Three Test Suites. EA-0 is zero-shot, EA-5 and EA-10 are 5-shot and 10-shot settings, respectively.}
\begin{center}
\begin{tabular}{|l|c|c|c|c|c|c|c|c|c|}
\hline
\textbf{Model} & \multicolumn{3}{c|}{\textbf{Single-line Bash}} & \multicolumn{3}{c|}{\textbf{Multiple-line Bash}} & \multicolumn{3}{c|}{\textbf{PowerShell}} \\
\hline
 & \textbf{EA-0} & \textbf{EA-5} & \textbf{EA-10} & \textbf{EA-0} & \textbf{EA-5} & \textbf{EA-10} & \textbf{EA-0} & \textbf{EA-5} & \textbf{EA-10} \\ 
\hline
GPT-4o & \textbf{84\%} & \textbf{86\%} & \textbf{88\%} & \textbf{68\%} & \textbf{68\%} & \textbf{66\%} & \textbf{64\%} & 64\% & \textbf{72\%} \\
\hline
CodeLlama-34b-inst & 70\% & 68\% & 64\% & 12\% & 4\% & 12\% & 52\% & \textbf{68\%} & 36\% \\
\hline
Mistral-7B-inst & 62\% & 54\% & 58\% & 12\% & 8\% & 2\% & 32\% & 28\% & 24\% \\
\hline
Mixtral-8x7B-inst & 70\% & 54\% & 32\% & 22\% & 8\% & 10\% & 52\% & 60\% & 44\% \\
\hline
Granite-20b-code-inst & 64\% & 64\% & 74\% & 26\% & 14\% & 8\% & 52\% & 56\% & 52\% \\
\hline
Granite-34b-code-inst & 62\% & 64\% & 62\% & 32\% & 16\% & 16\% & 44\% & 52\% & 64\% \\
\hline
Granite-34b-code-base & 76\% & \textbf{\textit{82\%}} & \textbf{\textit{84\%}} & 16\% & 6\% & 4\% & 48\% & 52\% & 52\% \\
\hline
\end{tabular}
\end{center}
\label{tab:evaluations}
\end{table*}

\begin{table*}[htbp]
\caption{Execution-based Accuracy between System administration (9/50 tasks) vs. Math problems (41/50 tasks) in Multiple-line Bash scripts. EA-0 is zero-shot, EA-5 and EA-10 are 5-shot and 10-shot settings, respectively.}
\begin{center}
\begin{tabular}{|l|c|c|c|c|c|c|}
\hline
\textbf{Model} & \multicolumn{2}{c|}{\textbf{EA-0}} & \multicolumn{2}{c|}{\textbf{EA-5}} & \multicolumn{2}{c|}{\textbf{EA-10}} \\
\hline
 & \textbf{System} & \textbf{Math} & \textbf{System} & \textbf{Math} & \textbf{System} & \textbf{Math} \\ 
\hline
GPT-4o & 33\% & \textbf{76\%} & \textbf{56\%} & \textbf{71\%} & \textbf{67\%} & \textbf{66\%}  \\
\hline
CodeLlama-34b-inst & 56\% & 2\% & 11\% & 2\% & 22\% & 10\%  \\
\hline
Mistral-7B-inst & 33\% & 7\% & 11\% & 7\% & 11\% & 0\% \\
\hline
Mixtral-8x7B-inst & \textbf{67}\% & 12\% & 22\% & 5\% & 11\% & 10\% \\
\hline
Granite-20b-code-inst & 22\% & 27\% & 11\% & 15\% & 33\% & 2\% \\
\hline
Granite-34b-code-inst & 33\% & 32\% & 22\% & 15\% & 22\% & 15\% \\
\hline
Granite-34b-code-base & 22\% & 15\% & 0\% & 7\% & 11\% & 2\% \\
\hline
\end{tabular}
\end{center}
\label{tab:evaluations_analysis_bash_scripts}
\end{table*}

\section{Error Analysis - Advantages and Challenges of Execution-based Evaluation}
Through error analysis (Sections \ref{sub:error1} and \ref{sub:error2}), execution-based evaluation proves its value with commands that either appeared to be correct, but weren't, or, conversely, appeared to be erroneous but were valid. Without execution-based evaluation, cases in Section \ref{sub:error1} would obtain high surface-form similarity score regardless they are not correct, whereas cases in \ref{sub:error2} would be considered low similarity score despite being correct results. This makes execution-based evaluation better, more realistic and focused on assessing the actual semantic correctness and functionality of generated codes than surface-form similarity metrics. Furthermore, Section \ref{sub:challenges} also presents challenges in using execution-based evaluation that we can improve in future.

\subsection{Bash command appears correct yet it is incorrect}
\label{sub:error1}
\begin{itemize}
    \item \textbf{Prompt:} What is the bash command to list the *.dat files in the current directory in ascending order by size?\\
    \textbf{Pred:} \texttt{ls -lh *.dat | sort -k5}\\
    \textbf{Result:} FAIL, size is not sorted correctly because \texttt{"ls -h"} outputs sizes in "human" units of K, eg, "1.1K"

    \item \textbf{Prompt:} What is the bash command to count the total number of lines in the *.c files under my src directory?\\
    \textbf{Pred:} \texttt{find src -type f -name "*.c" -exec wc -l {} + | awk '{sum += \$1} END {print sum}'}\\
    \textbf{Result:} FAIL, it produces 2 times of count because \texttt{"wc -l"} outputs it own total which \texttt{awk} included in its sum.

\end{itemize}

\subsection{Alternative solution appears different from ground-truth yet it is correct}
\label{sub:error2}
\begin{itemize}
    \item \textbf{Prompt:} What is the bash command to create a new file called data.dat with a size of 512K bytes?\\
    \textbf{Pred:} \texttt{touch data.dat; dd if=/dev/zero of=data.dat bs=512 count=1024}\\
    \textbf{Result:} Even though it included a spurious command, this was still a valid script.

    \item \textbf{Prompt:} What is the bash command to count all of the directories under my home directory?\\
    \textbf{Pred:} \texttt{find $\sim$ -type d -exec echo x; | wc -l}\\
    \textbf{Result:} \texttt{"-exec echo x"} is superfluous but the command still does the job.

    \item \textbf{Prompt:} What is the bash command to prefix the line number to every line in file.log?\\
    \textbf{Pred:} \texttt{sed = file.log | sed 'N;s/\textbackslash n/ /' | sed 's/\string^/echo -n "\string&"/e'}\\
    \textbf{Result:} Even though the last stage of the pipe is, effectively, a no-op, it satisfies the prompt.

    \item \textbf{Prompt:} What is the bash command to create a new file called data.dat with a size of 512K bytes?\\
    \textbf{Pred1:} \texttt{truncate -s 512k data.dat} \\
    \textbf{Pred2:} \texttt{dd if=/dev/zero of=data.dat bs=512 count=1024} \\
    \textbf{Pred3:} \texttt{fallocate -l512K data.dat} \\
    \textbf{Pred4:} \texttt{head -c 512K /dev/zero > data.dat} \\
    \textbf{Pred5:} \texttt{tail -c 512K /dev/zero > data.dat} \\
    \textbf{Result:} All of these Bash commands are correct.
\end{itemize}

\subsection{Challenges}
\label{sub:challenges}
Sometimes the generated command is correct but it proves too challenging for our evaluator. Usually this is because the model has added extra, unsolicited, steps to a command such writing the output to a file or excessive reformatting using awk and cut. Some examples:
\begin{itemize}
    \item \textbf{Prompt:} What is the bash command to prefix the line number to every line in file.log?\\
    \textbf{Pred:} \texttt{sed = file.log | sed 'N;s/\textbackslash n/ /' | sed 's/\string^/ /' > file.log.numbered}\\
    \textbf{Result:} Should PASS but marked FAIL because the output was sent to a file.

    \item \textbf{Prompt:} What is the bash command which will give me the date 90 days from today?\\
    \textbf{Pred:} \texttt{DATE=\$(date -d "+90 days" "+\%Y-\%m-\%d")}\\
    \textbf{Result:} Should PASS but marked FAIL because the result went to shell variable.
\end{itemize}

\textit{Math problems:} Another great challenge is generating multiple-line Bash scripts to solve math problems that we analyzed in Evaluations section (Table \ref{tab:evaluations_analysis_bash_scripts}). To investigate this, we selected some math problems from the multiple-line Bash prompts where most models failed, then we rephrased the questions to see whether these models can understand them (i.e. Prompt\#11 "\textit{what is the greatest common divisor of two numbers?}" or "\textit{how to find the greatest common divisor of two integers?}"). Our observation shows that despite understanding the given math problems and even being able to solve them (in one or some popular programming languages), most models (except GPT-4o) struggle to generate correct Bash scripts to implement the solution. This shows that from understanding a given math problem and actually being able to solve it, to generating correct Bash script to solve it is a big gap that these LLMs cannot bridge yet. Perhaps this can be improved in future models.

\section{Conclusions and Future Work}
We present a methodology to design and implement an execution-based evaluation platform for Bash and PowerShell codes. We created three test suites (125 hand-crafted test cases) and benchmark seven popular LLMs using our platform. We conduct error analysis and describe advantages and challenges of execution-based evaluation, especially for Bash and PowerShell. Unlike surface form similarity metrics, execution-based evaluation provides a more comprehensive and robust environment to test alternative Bash and PowerShell codes for the same given prompts. This approach also provides a transparent explanation why a Bash command appears almost correct but may fail despite of having very high score on exact/partial match metrics. This work not only focuses on evaluating Bash and PowerShell codes to select best model for incident remediation in APM, but also can extend to more generic test-cases to evaluate Bash and PowerShell in general. For future work, we will refine and expand our test cases to cover more challenging tasks.

\appendix
\section{Appendices}
\label{sec:appendix}
We share this Execution-based benchmark consisting of test cases and podman images, environment settings, scripts, verifiers in this repository\footnote{https://github.com/IBM/nl2bash-eabench/tree/main}. We also provide an end-to-end example of execution-based verification (Appendix \ref{subsec:verifier}) for better understanding the complexity of this work.

\subsection{An End-to-end Example of Execution-based Verification}
\label{subsec:verifier}
We present an end-to-end example of execution-based verification.

\textbf{Prompt:} What is the bash command to list the *.dat files in the current directory in ascending order by size?

\textbf{Step-by-step process to evaluate the generated Bash code:}

\textit{\textbf{Step 1:}} The directory \texttt{/home/test} is created for the Podman container and populated with:
\begin{itemize}
    \item \texttt{Test.sh} – this is a driver which executes the following scripts in order:
    \item \texttt{Pre\_test.sh} – any "warm-up" needed; for this test it creates the \texttt{*.dat} files
    \item \texttt{Bash.sh} – the command from the LLM to be tested
    \item \texttt{Post\_test.sh} – any "in-containter" testing that may be required; In this test, it just lists the \texttt{*.dat} files
\end{itemize}

\textit{\textbf{Step 2:}} The test container runs and executes "test.sh"
\begin{itemize}
    \item \texttt{pre-test.sh}: the \texttt{*.dat} are created in random order so they are not "naturally" in order of by size
        \begin{itemize}
            \item fallocate -l 1030 \texttt{medium.dat}
            \item fallocate -l 1050 \texttt{grand.dat}
            \item fallocate -l 1010 \texttt{tiny.dat}
            \item fallocate -l 1040 \texttt{large.dat}
            \item fallocate -l 1020 \texttt{small.dat}
            \item fallocate -l 1060 \texttt{super.dat}
        \end{itemize}
    \item \texttt{bash.sh}, containing the generated command, is executed.
    \item \texttt{post-test.sh}: the \texttt{*.dat} files are tagged with "POST" for identification and listed on the standard output from the Podman container.
        \begin{itemize}
            \item \# Check for "naturally" ordered by size
            \item find . -name "*.dat" | sed -e "s+\^+POST: +"
            \item POST: ./small.dat
            \item POST: ./medium.dat
            \item POST: ./tiny.dat
            \item POST: ./large.dat
            \item POST: ./super.dat
            \item POST: ./grand.dat
        \end{itemize}
\end{itemize}

\textit{\textbf{Step 3:}} After the container exits, an evaluation script examines the Podman log file (Table \ref{tab-verification-script-1} \& \ref{tab-verification-script-2}) for final result.

\begin{table*}[htbp]
\caption{A Verification Script - PART 1.}
\begin{center}
\begin{tabular}{|l|}
\hline
\begin{lstlisting}[language=bash]
##!/bin/bash

# Look for any stderr output during runtime
rc=$(strip_out_stderr.sh podman.log | wc -l)
if [ $rc != 0 ]
then
    strip_out_stderr.sh podman.log >&2
    exit 2
fi

# No files should've been created
if [ $(stat --format="%s" diff.log) != 0 ]
then
    echo >&2 "Created spurious system files in the container -- see diff.log"
    exit 2
fi

# Only *.dat files should appear - check for the script files
# which are in the same directory
(strip_out_stdout.sh podman.log | grep "pre_test.sh" > /dev/null) || \
(strip_out_stdout.sh podman.log | grep "post_test.sh" > /dev/null) || \
(strip_out_stdout.sh podman.log | grep "test.sh" > /dev/null) || \
(strip_out_stdout.sh podman.log | grep "bash.sh" > /dev/null)
if [ $? == 0 ]
then
    echo >&2 "Only *.dat files should appear in stdout"
    exit 2
fi

# Extract the lines marked with "POST" on stdout
strip_out_stdout.sh podman.log | sed -e '/\.dat/!d' | \
sed -e "s+^STDOUT: ++" | sed -e "/^POST: /!d" | nl > tmp.log

# This returns "true" if the files are "naturally" in order
(grep -E "^ *1.*tiny.dat"   tmp.log > /dev/null) && \
(grep -E "^ *2.*small.dat"  tmp.log > /dev/null) && \
(grep -E "^ *3.*medium.dat" tmp.log > /dev/null) && \
(grep -E "^ *4.*large.dat"  tmp.log > /dev/null) && \
(grep -E "^ *5.*grand.dat"  tmp.log > /dev/null) && \
(grep -E "^ *6.*super.dat"  tmp.log > /dev/null)
no=$?

# This time ignore the POSTed lines to extract the output from the command generated by the LLM
strip_out_stdout.sh podman.log | sed -e '/\.dat/!d' | sed -e "s+^STDOUT: ++" \
| sed -e "/^POST: /d" | nl > tmp.log

# Are all of the files on a single line?
rc=$(wc -l tmp.log | sed -e 's/ *\([1-9][0-9]*\).*/\1/')
if [ $rc == 1 ]
then
    # There's just one line - is it sorted?
    grep -E "tiny.dat.*small.dat.*medium.dat.*large.dat.*grand.dat.*super.dat" tmp.log > /dev/null
    if [ $? == 0 ]
    then
        # Sorted
        if [ $no == 0 ]
        then
            # No choice but to accept it - but issue a warning that this may have been an accident
            echo "(You may have gotten lucky: they are \"naturally\" in order by size)"
        fi
        exit 0
    fi

    # Wasn't ascending so check for descending
    grep -E "super.dat.*grand.dat.*large.dat.*medium.dat.*small.dat.*tiny.dat" tmp.log > /dev/null
    if [ $? == 0 ]
    then
        # It's a "fail" but output the information
        echo >&2 "Appears to have sorted in descending rather than ascending order"
        exit 2
    fi

    # A single, unsorted, line - fail
    echo >&2 "Doesn't appear to have sorted the files by size"
    exit 2
fi

\end{lstlisting}
\end{tabular}
\end{center}
\label{tab-verification-script-1}
\end{table*}

\begin{table*}[htbp]
\caption{A Verification Script - PART 2.}
\begin{center}
\begin{tabular}{|l|}
\begin{lstlisting}[language=bash]
# More than one line output to stdout
# Check if sorted in ascending order
(grep -E "^ *1.*tiny.dat"   tmp.log > /dev/null) && \
(grep -E "^ *2.*small.dat"  tmp.log > /dev/null) && \
(grep -E "^ *3.*medium.dat" tmp.log > /dev/null) && \
(grep -E "^ *4.*large.dat"  tmp.log > /dev/null) && \
(grep -E "^ *5.*grand.dat"  tmp.log > /dev/null) && \
(grep -E "^ *6.*super.dat"  tmp.log > /dev/null)
if [ $? == 0 ]
then
    # ... but was it an accident?
    if [ $no == 0 ]
    then
        # Have to accept the result but issue a warning
        echo "(You may have gotten lucky: they are \"naturally\" in order by size)"
    fi
    exit 0
fi

# At this point, the output is wrong
# Is it sorted in descending order?
(grep -E "^ *6.*tiny.dat"   tmp.log > /dev/null) && \
(grep -E "^ *5.*small.dat"  tmp.log > /dev/null) && \
(grep -E "^ *4.*medium.dat" tmp.log > /dev/null) && \
(grep -E "^ *3.*large.dat"  tmp.log > /dev/null) && \
(grep -E "^ *2.*grand.dat"  tmp.log > /dev/null) && \
(grep -E "^ *1.*super.dat"  tmp.log > /dev/null)
if [ $? == 0 ]
then
    echo >&2 "Expected ascending rather than descending order"
    exit 2
fi

# Not sorted
echo >&2 "Doesn't appear to have sorted the files by size"

# Some generated commands attempted to sort the files after expressing their sizes in units of KB
# (File sizes were deliberately chosen to be around 1KB)
grep -E "[0-9]K " tmp.log > /dev/null
if [ $? == 0 ]
then
    echo >&2 "Was the filesize output in units of \"K\"?"
fi

exit 2

\end{lstlisting}
\end{tabular}
\hrule
\end{center}
\label{tab-verification-script-2}
\end{table*}





\newpage
\pagebreak

\end{document}